\documentclass{article}

\usepackage[accepted]{icml2014}
\usepackage{times}
\usepackage{graphicx}
\usepackage{natbib}
\usepackage{hyperref}
\usepackage{subfigure}
\usepackage{algorithm}
\usepackage{algorithmic}

\newcommand{\ignore}[1]{}

\icmltitlerunning{Teaching Deep Convolutional Neural Networks to Play Go}

\begin{document}

\twocolumn[
\icmltitle{Teaching Deep Convolutional Neural Networks to Play Go}

\icmlauthor{Christopher Clark}{s1351418@sms.ed.ac.uk}
\icmladdress{University of Edinburgh}
\icmlauthor{Amos Storkey}{a.storkey@ed.ac.uk}
\icmladdress{University of Edinburgh}


\vskip 0.3in
]

\begin{abstract}
Mastering the game of Go has remained a long standing challenge to the field of AI. Modern computer Go systems rely on processing millions of possible future positions to play well, but intuitively a stronger and more `humanlike' way to play the game would be to rely on pattern recognition abilities rather then brute force computation. Following this sentiment, we train deep convolutional neural networks to play Go by training them to predict the moves made by expert Go players. To solve this problem we introduce a number of novel techniques, including a method of tying weights in the network to `hard code' symmetries that are expect to exist in the target function, and demonstrate in an ablation study they considerably improve performance. Our final networks are able to achieve move prediction accuracies of 41.1\% and 44.4\% on two different Go datasets, surpassing previous state of the art on this task by significant margins. Additionally, while previous move prediction programs have not yielded strong Go playing programs, we show that the networks trained in this work acquired high levels of skill. Our convolutional neural networks can consistently defeat the well known Go program GNU Go, indicating it is state of the art among programs that do not use Monte Carlo Tree Search. It is also able to win some games against state of the art Go playing program Fuego while using a fraction of the play time. This success at playing Go indicates high level principles of the game were learned.\end{abstract}

\section{Introduction}
Go is an ancient, deeply strategic board game that is notable for being one of the few board games where human experts are still comfortably ahead of computer programs in terms of skill. Predicting the moves made by expert players is an interesting and challenging machine learning task, and has immediate applications to computer Go. In this section we provide a brief overview of Go, previous work, and the motivation for our deep learning based approach.

\subsection{The Game of Go}
\begin{figure}[ht]
\begin{center}
\includegraphics[scale=0.5]{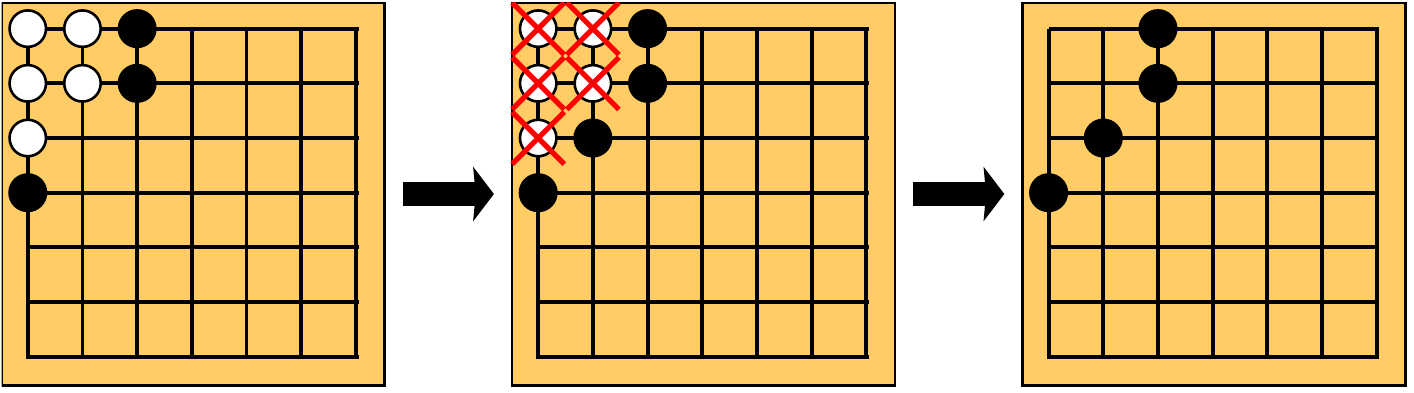}
\caption{Example of capturing pieces in Go. Here white's stones in the upper left are connected to each other through adjacency so they form a single group (left panel). When black places a stone on the indicated grid point (middle panel) that group is surrounded, meaning there are no longer any empty grid points adjacent to it, so the entire group is removed from the board (right panel).}
\label{go_capturing}
\end{center}
\end{figure}

\begin{figure}[ht]
\begin{center}
\subfigure{
	\includegraphics[scale=0.35]{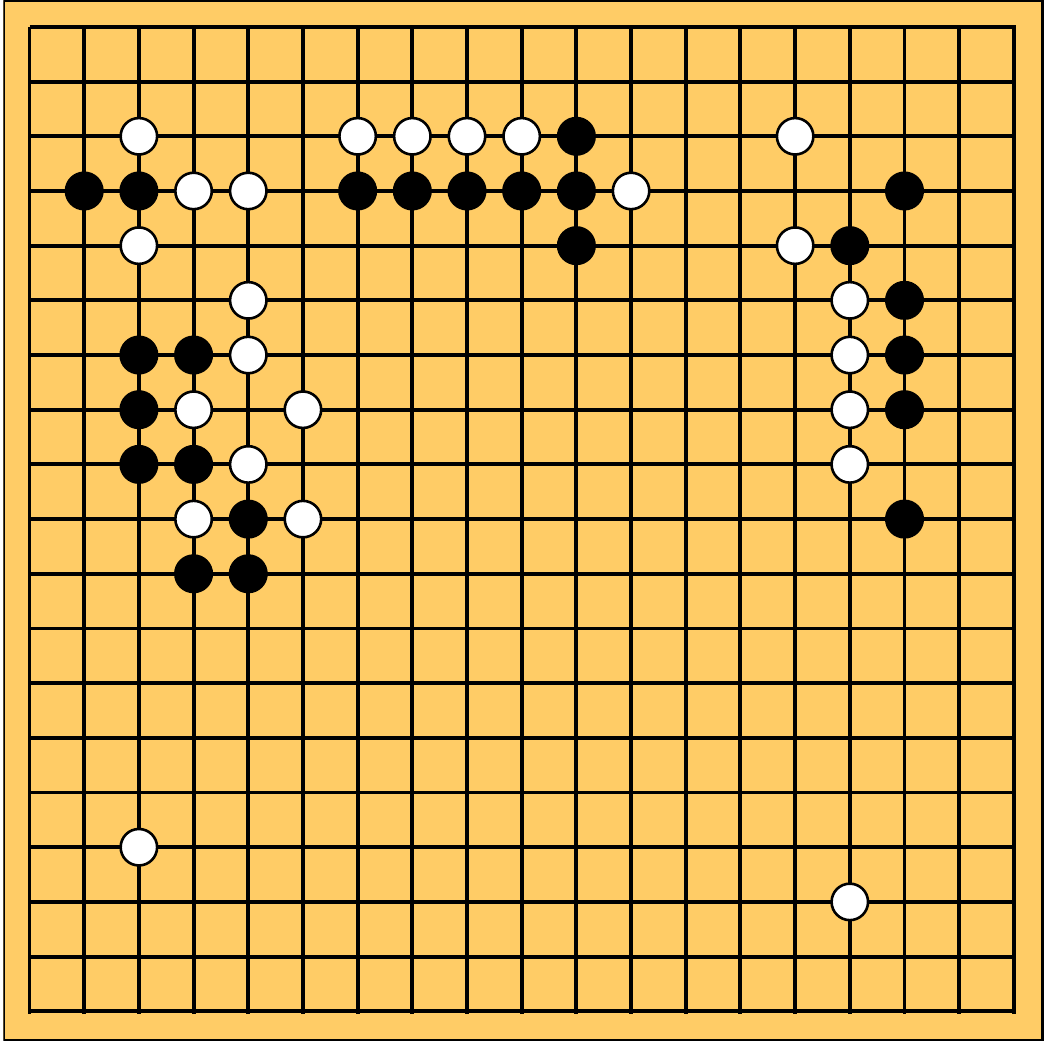}
}
\subfigure{
	\includegraphics[scale=0.35]{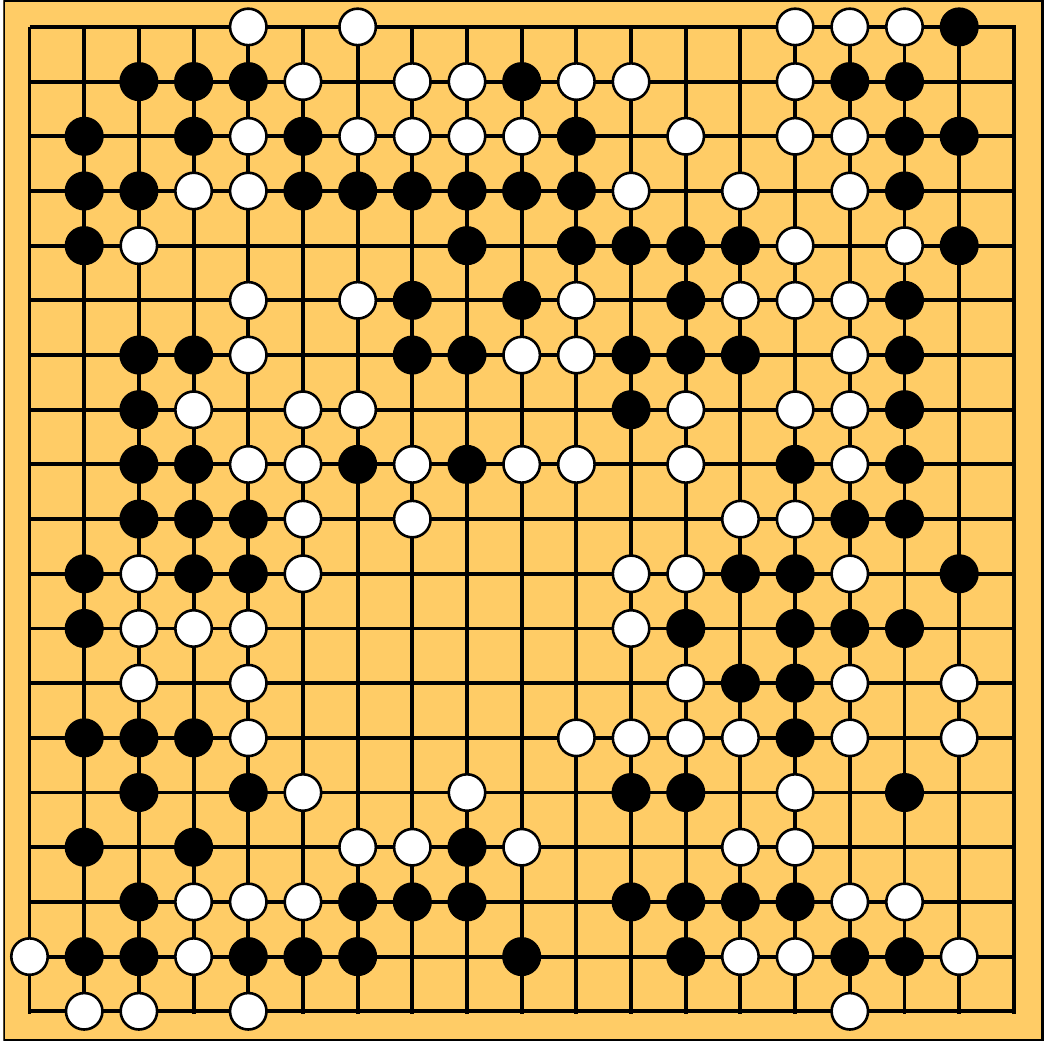}
}
\caption{Example of positions from a game of Go after 50 moves have passed (left) and after 200 moves have passed (right). In the right panel it can be seen that white is gaining control of territory in the center and top of the board, while black is gaining influence over the left and right edges.}
\label{go_example}
\end{center}
\end{figure}

We give a very brief introduction to the rules of Go. We defer to~\cite{bozulich1992go} or~\cite{muller_computer_go} for a more comprehensive account of the rules. Go has a number of different rulesets that subtly differ as to when moves are illegal and how the game is scored, here we focus on generalities that are common to all rulesets.

Go is a two player game that is usually played on a 19x19 grid based board. The board typically starts empty, but the game can start with stones pre-placed on the board to give one player a starting advantage. Black plays first by placing a black stone on any grid point he or she chooses. White then places a white stone on an unoccupied grid point and play continues in this manner. Players can opt to pass instead of placing a stone, in which case their turn is skipped and their opponent may make a second move. Stones cannot be moved once they are placed, however a player can capture a group of their opponent's stones by surrounding that group with their own stones. In this case the surrounded group is removed from the board as shown in Figure~\ref{go_capturing}. Broadly speaking, the objective of Go is to capture as many of the grid points on the board as possible by either occupying them or surrounding them with stones. The game is played until both players pass their turn, in which case the players come to an agreement about which player has control over which grid points and the game is scored.

Through the capturing mechanic it is possible to create infinite `loops' in play as players repeatedly capture and replace the same pieces. Go rulesets include rules to prevent this from occurring. The simplest version of this rule is called the simple-ko rule, which states that players cannot play moves that would recreate the position that existed on their previous turn. Most Go rulesets contain stronger versions of this rule called super-ko rules, which prevent players from recreating any previously seen position. Figure~\ref{go_example} shows some example board positions that could occur as a game of Go is played.

State of the art computer Go programs such as Fuego~\cite{fuego}, Pachi~\cite{pachi}, or CrazyStone~\footnote{http://remi.coulom.free.fr/CrazyStone/}, can achieve the skill level of very strong amateur players, but are still behind professional level play. The difficulty computers have in this domain relative to other board games, such as chess, is often attributed to two things. First, in Go there are a very large number of possible moves. Players have $19\times19=361$ possible starting moves. As the board fills up the number of possible moves decreases, but can be expected to remain in the hundreds until late in the game. This is in contrast to chess where the number of possible moves might stay around fifty. Second, good heuristics for assessing which player is winning in Go have not been found. In chess, examining which pieces each player has left on the board, plus some heuristics to assess which player has a better position, can serve as a good estimator for who is winning. In Go counting the number of stones each player has is a poor indicator of who is winning, and it has proven to be difficult to build effective rules for estimating which player has the stronger position.

Current state of the art Go playing programs use Monte Carlo Tree Search (MCTS) algorithms to tackle these difficulties. MCTS algorithms evaluate positions in Go using simulated `playouts' where the game is played to completion from the current position assuming both players move randomly or follow some cheap best move heuristic. Many playouts are carried out and it is then assumed good positions are ones where the program was the winner in the majority of them. Finding improvements to this strategy has seen a great deal of recent work and has been the main source of progress for building Go playing programs. See~\cite{browne2012survey_of_mcts_methods} for a recent survey of MCTS algorithms and~\cite{rimmel2010current_frontiers_in_computer_go} for a survey of some modern Go playing systems.

\subsection{Move Prediction in Go}
Human Go experts rely heavily on pattern recognition when playing Go. Expert players can gain strong intuitions about what parts of the board will fall under whose control and what are the best moves to consider at a glance, and without needing to mentally simulate possible future positions. This provides a sharp contrast to typical computer Go algorithms, which simulate thousands of possible future positions and make minimal use of pattern recognition. This gives us reason to think that developing pattern recognition algorithms for Go might be the missing element needed to close the performance gap between computers and humans. In particular for Go, pattern recognition systems could provide ways to combat the high branching factor because it might be possible to prune out many possible moves based on patterns in the current position. This would result in a system that analyzes Go in a much more `human like' manner, eliminating most candidate moves based on learned patterns within the board and then `thinking ahead' only for the remaining, more promising moves. In this work we train deep convolutional neural networks (DCNNs) to detect strong moves within a Go position by training them on move prediction, the task of predicting where expert human players would choose to move when faced with a particular position.

Outside of playing Go, move prediction is an interesting machine learning task in its own right. We expect the target function to be highly complex, since it is fair to assume human experts think in complex, non-linear ways when they choose moves. We also expect the target function to be non-smooth because a minor change to a position in Go (such as adding or removing a single stone) could be expected to dramatically alter which moves are likely to be played next. These properties make this learning task very challenging, however it has been argued that acquiring an ability to learn complex, non-smooth functions is of particular importance when it comes to solving AI~\cite{bengio2009learning}. These properties have also motivated us to attempt a deep learning approach, as it has been argued that deep learning is well suited to learning complex, non-smooth functions~\cite{bengio2009learning, bengio2007scaling}. Deep learning has allowed machine learning algorithms to approach human level pattern recognition abilities in image and speech recognition, this work shows this success can be extended to the more AI oriented domain of move prediction in Go.

\subsection{Previous Work}
Previous work in move prediction for Go typically made use of feature construction or shallow neural networks. The feature construction approach involves characterizing each legal move by a large number of features. These features include many `shape' features, which take on a value of 1 if the stones around the move in question match a predefined configuration of stones and 0 otherwise. Stone configurations can be as small as the nearest two or three squares and as large as the entire board. Very large numbers of stone configurations can be harvested by finding commonly occurring stone configurations in the training data. Shape features can be augmented with hand crafted features, such as distance of the move in question to the edge of the board, whether making the move will capture or lose stones, its distance to the previous moves, ect. Finally a model is trained to rank the legal moves based on their features. Work following this approach includes~\cite{stern2006bayesian_pattern_ranking_for_move_prediction,araki2007move_prediction_in_go_with_max_ent,wistuba2012comparison_bayesian_move_prediction_systems,wistuba2013move_prediction_modeling_feature_interaction,coulom2007computing_elo_of_move_patterns}. Depending on the complexity of the model used, researchers have seen accuracies of 30 - 41\% on move prediction for strong, amateur Go players. This best algorithm we know of achieved 41\%~\cite{wistuba2013move_prediction_modeling_feature_interaction} using features of this manner and a non-linear ranking model.

A few attempts to use neural networks for move prediction have been completed before. Work by~\cite{van2003local_move_prediction_in_go} trained a neural network to predict expert moves using a set of hand constructed features, preprocessing techniques to reduce the dimensionality of the data, and a two layer neural network that predicted whether a particular move was one a professional player would have made or not. The network was able to achieve 25\% accuracy on a test set of professional games. This work more closely resembles the work done by~\cite{sutskever2008mimicking_go_experts}, where two layer convolutional networks were trained for move prediction. Their networks typically included a convolution layer with fifteen 7x7 filters followed by another convolutional layer with one 5x5 filter. Each layer zero-padded its input so that its output had 19x19 shape. A softmax function was applied at the final layer and the output interpreted as the probabilities an expert player would place a stone on different grid points. They achieved an accuracy of 34\% using a network that took both the current board position and the previous moves as input. Using an ensemble the accuracy reached 37\%.

Our work will differ in several important ways. We use much deeper networks and propose several novel position encoding schemes and network designs that improve performance. We find that the most important one is a strategy of tying weights within the network to `hard code' particular symmetries that we expect to exist in good move prediction functions. We are also careful not to use the previously made moves as input. There are two motivations for choosing not to do so. First, classifiers trained using previous moves as input might come to rely on simple heuristics like `move near the area where previous moves were made' instead of learning to evaluate positions based on the current stone configuration. While this might improve accuracy, our fundamental motivation is to see whether the classifier can capture Go knowledge, and the ability to borrow knowledge from experts by looking at their past moves cheapens this objective. From a game theoretic perspective the previous moves should not influence what the current best move is, so this information should not be needed to play well. Secondly, using the previous move is likely to reduce performance when it comes to playing as a stand alone Go player. During play both an opponent and the network are liable to make much worse moves then would be made by Go experts, therefore coming to rely on the assumption that the previous moves were made by experts is likely to yield bad results. This potential problem was also noted by~\cite{araki2007move_prediction_in_go_with_max_ent}. This work is also the first one to perform an evaluation across two datasets, providing an opportunity to compare how classifiers trained on these datasets differ in terms of Go playing abilities and move prediction accuracy.

Several of the works mentioned above analyze or comment on the strength of their move predictor program as a stand alone Go player. In~\cite{van2003local_move_prediction_in_go} researchers found that their neural network was consistently defeated by GNU Go and conclude their `...belief was confirmed that the local move predictor in itself is not sufficient to play a strong game.' Work by~\cite{araki2007move_prediction_in_go_with_max_ent} also reports that their move predictor was beaten by GNU Go. Researchers in~\cite{stern2006bayesian_pattern_ranking_for_move_prediction} report that other Go players estimated their move predictor as having a ranking of 10-15 kyu, but do not report its win rates against another computer Go opponent. Both~\cite{coulom2007computing_elo_of_move_patterns, wistuba2013move_prediction_modeling_feature_interaction} do not give formal results, but state that their systems did not make strong stand alone Go playing programs. In general it seems that past approaches to move prediction have not resulted in Go programs with much skill.

\section{Approach}
\subsection{Data Representation}
As done by~\cite{sutskever2008mimicking_go_experts}, the networks trained here take as input a representation of the current position and output a probability distribution over all grid points in the Go board, which are interpreted as a probability distribution over the possible places an expert player could place a stone. During testing probability given to grid points that would be an illegal move, either due to being occupied by a stone or due to the simple-ko rule, are set to zero and the remaining outputs renormalized. We follow~\cite{sutskever2008mimicking_go_experts} by encoding the current position in two 19x19 binary matrices. The first matrix has ones indicating the location of the stones of the player who is about to play, the second 19x19 matrix has ones marking where the opponent's stones are. We depart from~\cite{sutskever2008mimicking_go_experts} by additionally encoding the presence of a `simple-ko constraint' if one is present in a third 19x19 matrix. Here simple-ko constraints refers to grid points that the current player is not allowed to place a stone on due to the simple-ko rule. Simple-ko constraints are reasonably sparse. In our dataset of professional games only 2.4\% of moves are made with a simple-ko constraint present. However simple-ko constraints are often featured in Go tactics so we hypothesize they are still important to include as input. We elect not to encode move constraints beyond the ones created by the simple-ko rule, meaning constraints stemming from super-ko rules, because they are rare, harder to detected, ruleset-dependent, and less prominent in Go tactics. Thus the input has three channels and a height and width of 19. Again following~\cite{sutskever2008mimicking_go_experts} as well as other work that has found this to be a useful feature such as~\cite{wistuba2013move_prediction_modeling_feature_interaction}, we tried encoding the board into 7 channels where stones are encoded based on the number of `liberties', or the number of empty squares that the opposing player would need to occupy to capture that stone. In this case channels 1-3 encode the current player's pieces with 1, 2, or 3 or more liberties, channels 4-6 do the same for the opponent's pieces, and the 7th channel marks simple-ko constraints as before.

The classifier is not trained to predict when players choose to pass their move. Passing is extremely rare throughout most of the game because it is practically never beneficial to pass a move in Go. Thus passing is mainly used at the end of the game to signal willingness to end the game. This means players usually pass, not because there are no beneficial moves left to be played, but due to being in a situation where both players can agree the game is over. Modeling when this situation occurs is beyond the scope of this work.

\subsection{Basic Architecture}
Our most effective networks were composed of many convolutional layers. Since the input is only of size 19x19 we found it important to zero pad the input to each convolutional layer to stop the size of the outputs of the higher layers becoming exceedingly small. We briefly experimented with some variations, but found that zero-padding each layer to ensure each layer's output has a 19x19 shape was a reasonable choice. In general using a fully connected top layer was more effective than a convolutional top layer as was used by~\cite{sutskever2008mimicking_go_experts}. However, we found the performance gap between networks using a convolutional and fully connected top layer decreased as the networks increased in size. As the networks increased in size using more then one fully connected layer at the top of the network became unhelpful. Thus our architectures use many convolutional layers followed by a single fully connected top layer.

In our experience using the rectifier activation function was slightly more effective then using the tanh activation function. We were limited primarily by running time, in almost all cases increasing the depth and number of filters of the network increased accuracy. This implies we have not hit the limit of what can be achieved merely by scaling up the network. We found that using many, smaller convolutional filters and deep networks was more efficient in terms of trading off runtime and accuracy then using fewer, larger filters. We briefly experimented with non-convolutional networks but found them to be much harder to train, often requiring more epochs of training and the use of approximate second order gradient descent methods, while getting worse results. This makes us think that a convolutional architecture is important for doing well in this task.

\subsection{Additional Design Features}
Along with the network architecture described above we introduce a number of novel techniques that we found to be effective in this domain.
\subsubsection{Edge Encoding}
In the neural networks described so far the first convolutional layer will zero pad its input. In the current board representation zeros represent empty grid points, so this zero padding results in the board `appearing' to be surrounded by empty grid points. In other words, the first layer cannot capture differences between stones being next to an edge or an empty grid point. A solution is to reserve an additional channel to encode the out of bounds squares of the padded input. In this scheme a completely empty eighth channel is added to the input. Then, instead of zero padding the input, the input is padded with ones around the new eighth channel and padded with zeros around the other channels. This allows the network to distinguish out of bound squares from empty ones. We experimented with padding the input with ones in the channel used for the opponent's pieces and zeros in the other channels, but found this to give worse results.
\subsubsection{Masked Training}
In Go some grid points of the board are not legal moves, either because they are already occupied by a stone or due to ko rules. Therefore these points can be eliminated as possible places an expert will move a priori. During testing we account for this, but this knowledge is not present in the network during training. Informal experiments show that the classifier is able to learn to avoid predicting illegal moves with very close to 100\% accuracy, but we still speculate that accounting for this knowledge during training might be beneficial because it will simplify the function the classifier is trying to learn. To accomplish this we zero out the outputs from the top layer that are illegal, and then apply the softmax operator, make predictions, and backprop the gradient across only these outputs during learning.

\subsubsection{Enforcing Reflectional Preservation}
\begin{figure}[ht]
\begin{center}
\subfigure{
	\includegraphics[scale=0.45]{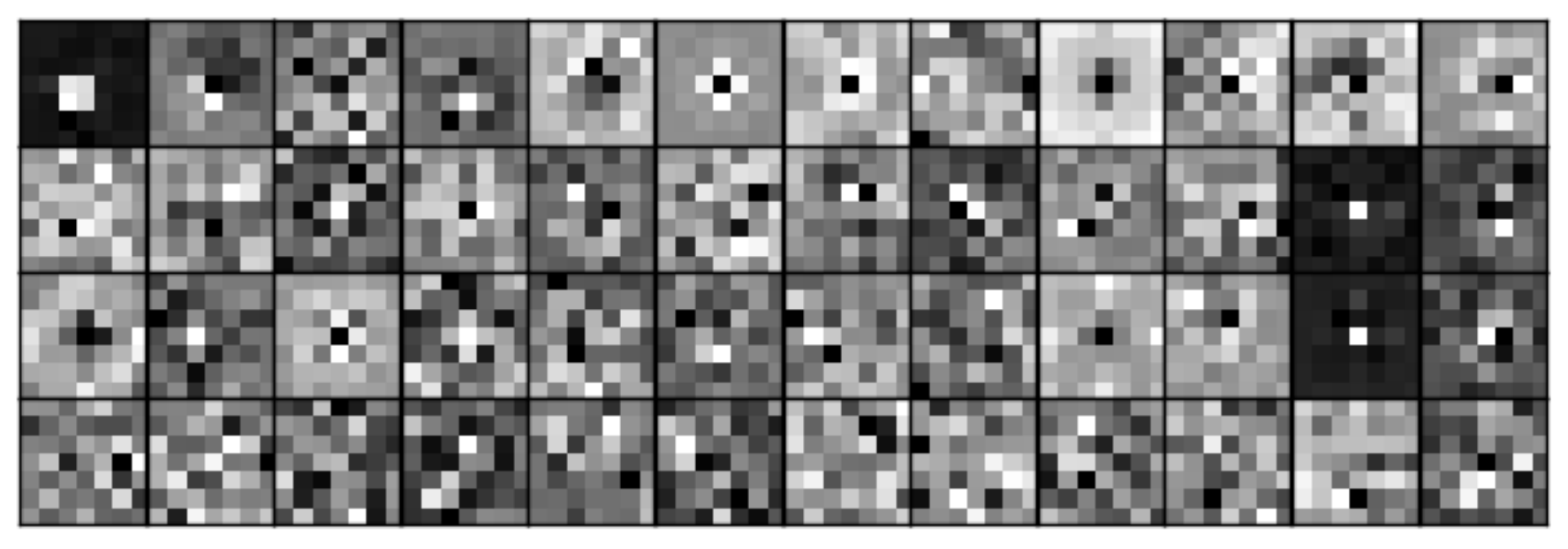}
}
\subfigure{
	\includegraphics[scale=0.45]{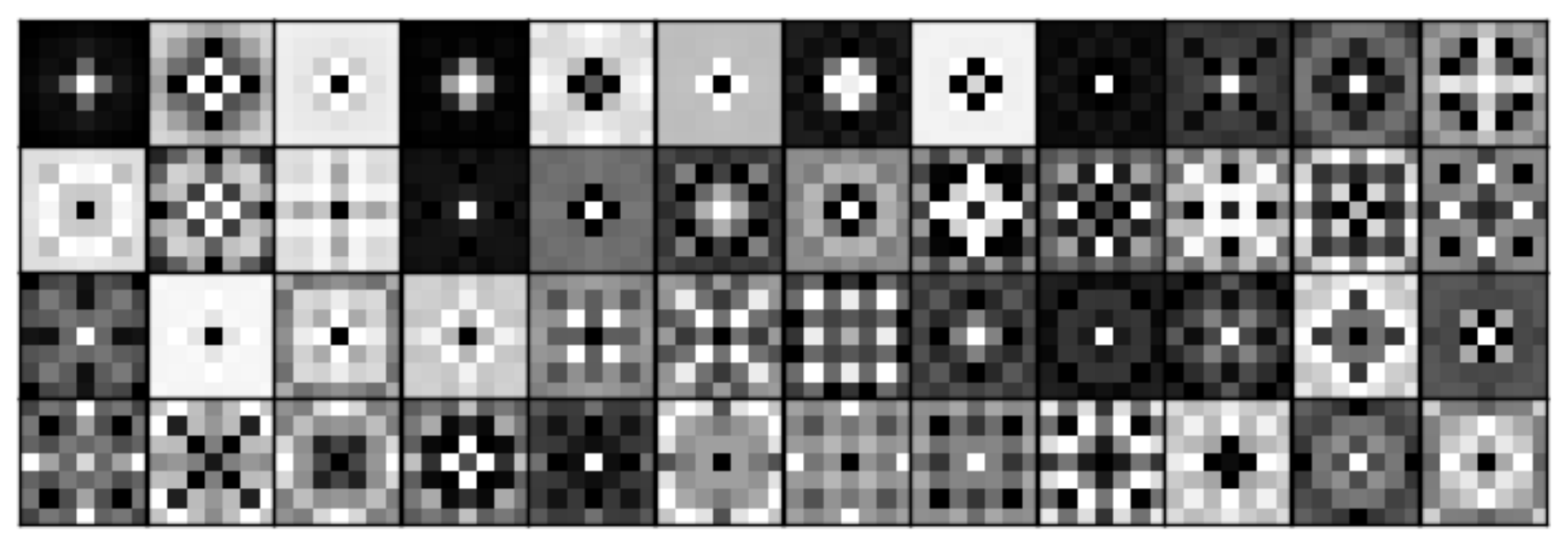}
}
\caption{Visualization of the weights of randomly sampled channels from randomly sampled convolution filters of a five layer convolutional neural network trained on the GoGoD dataset. The network was trained once without (top) and once with (bottom) reflection preservation being enforced. It can be seen that even without weight tying some filters, such as row 1 column 6, have learned to acquire a symmetric property. This effect is even stronger in the weight visualization in~\cite{sutskever2008mimicking_go_experts}.}\label{convolutional_weights}.
\end{center}
\end{figure}

\begin{figure}[ht]
\begin{center}
\includegraphics[scale=0.60]{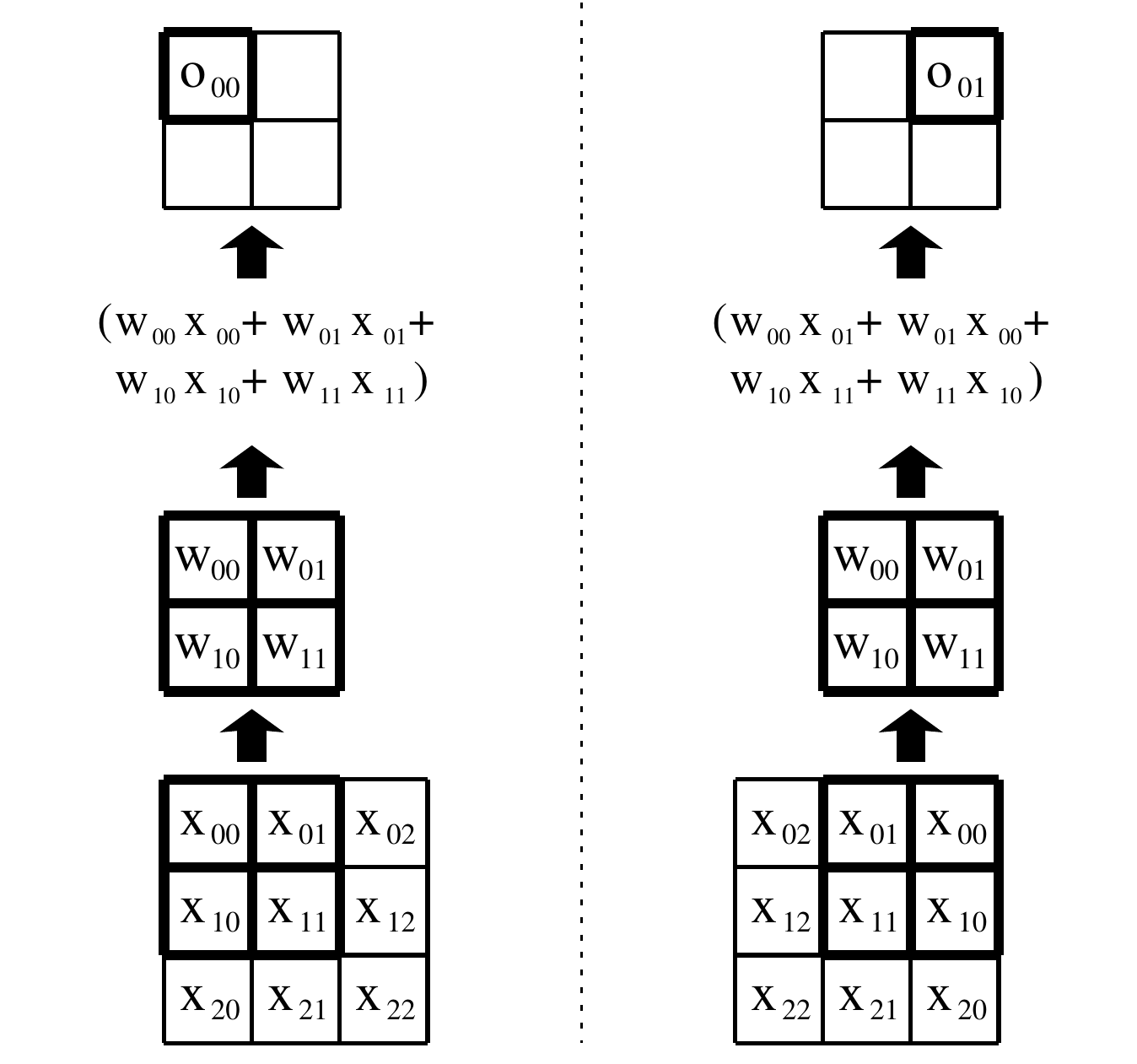}
\caption{Applying the same convolutional filter to the upper left patch of an input (left) and to the upper right patch of the same input reflected across its y-axis (right). To hard encode horizontal reflection preservation we will need to ensure $o_{00} = o_{01}$, for any $x$. This means we must have $w_{00} = w_{01}$ and $w_{10} = w_{11}$.}
\label{conv_invar_graphic}
\end{center}
\end{figure}

\begin{figure}[ht]
\begin{center}
\includegraphics[scale=0.60]{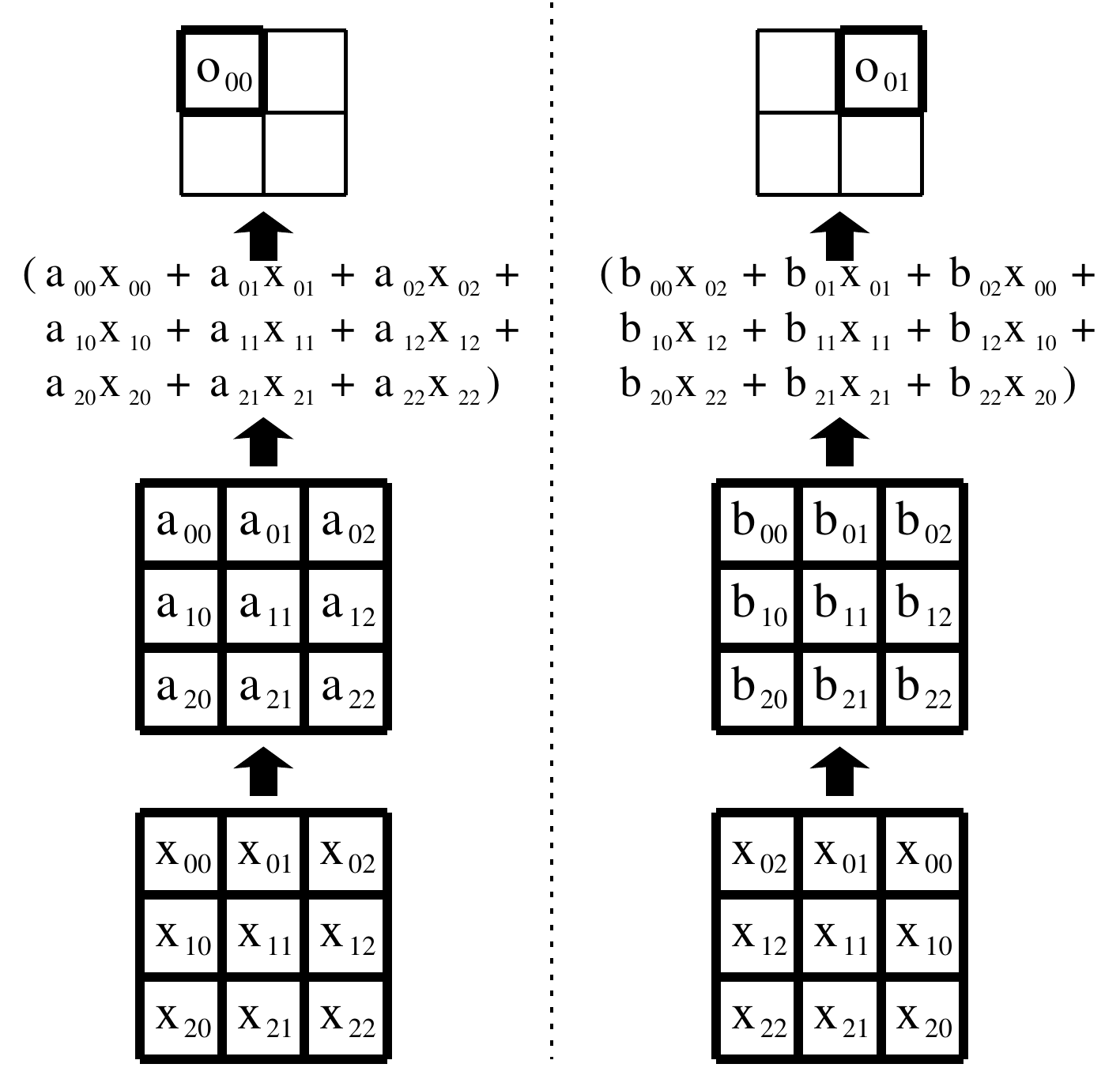}
\caption{The computation for a single output unit, $o_{00}$ with weights $a_{ij}$, for an input (left) and the computation for the output unit on the opposite side of the vertical axis, $o_{01}$ with weights $b_{ij}$ when the input is reflected across the y-axis. To hard encode horizontal reflection preservation we need to ensure $o_{00} = o_{01}$ for any $x$. It is easy to see by examining the equations shown that this means we require $a_{00} = b_{02}$, $a_{01} = b_{01}$, $a_{02} = b_{00}$, ect.}
\label{fc_invar_graphic}
\end{center}
\end{figure}

In Go, if the board is reflected across either the x, y, or diagonal axis the game in some sense has not changed. The transformed position could be played out in an identical manner as the original position (relative to the transformation), and would result in the same scores for each player. Therefore we would expect that, if the Go board was reflected, the classifier's output should be reflected in the same manner. One way of exploiting this insight would be to train on various combinations of reflections of our training data, increasing the number of data points we could train on by eight folds. This tactic comes at a serious cost, our final network took four days to train for ten epochs. Increasing our data by eight folds means it would require almost a month to train for the same number of epochs.

A better route is to `hard wire' this reflectional preserving property into the network. This can be done by carefully tying the weights so that this property exists for each layer. For convolutional layers, this means tying the weights in each convolutional filter so that they are invariant to reflections. In other words enforcing that the weights are symmetrical around the horizontal, vertical, and diagonal dividing lines. An illustration of the kinds of weights this produces can be found in Figure~\ref{convolutional_weights}. To see why this has the desired effect consider the application of a convolutional filter to an input with one channel. Let $w_{ij}$ be the weights of the convolutional filter and $x_{ij}$ index into any square patch from the input where $0 <= i < n$ and $0 <= j < m$ and $i$ and $j$ index into the row and column of the input patch/weight in a top-to-bottom and right-to-left manner. The pre-activation output of this filter when applied to this patch of input is $s = \sum\limits{_{i=0}^{n - 1}} \sum\limits{_{j=0}^{m - 1}} x_{ij}w_{ij}$. Should the input be reflected horizontally, the patch of inputs $x_{ij}$ will now be reflected and located on the opposite side of the vertical axis. Assuming the convolutional filter is applied in a dense manner (a stride of size 1), the same convolutional filter will be applied to the reflected patch. It is necessary and sufficient that the application of the convolutional filter to this reflected patch has a pre-activation value of $s$, the same as before, to meet our goal of having this layer preserve horizontal reflections applied to the input. Thus we want $\sum\limits{_{i=0}^{n - 1}} \sum\limits{_{j=0}^{m - 1}} x_{ij}w_{ij} = \sum\limits{_{i=0}^{n - 1}} \sum\limits{_{j=0}^{m - 1}} x_{i{(m - j)}}w_{ij}$ for any  $x_{ij}$. Thus we require $w_{ij} = w_{i(m -j)}$, in short that the convolutional filter is symmetric around the vertical axis. This is illustrated in Figure~\ref{conv_invar_graphic}. A similar proof can be built for vertical or diagonal reflections.

For fully connected layers this property can also be encoded using weight tying. First, we arrange the outputs of the layer into a square shape. Then let $a$ be any output unit and $a_{ij}$ refer to the weight given to input $x_{ij}$ for that unit. Let $b$ be the output unit that is the horizontal reflection of $a$ and $b_{ij}$ refer to the weight given to input $x_{ij}$ for output $b$. Let $s$ be the pre-activation value for $a$, then  $s = \sum\limits{_{i=0}^{n - 1}} \sum\limits{_{j=0}^{m - 1}} {a_{ij} x_{ij}}$ where $m$ and $n$ are the height and width of the input, also assumed to have a square shape. To achieve invariance to horizontal reflections we require the pre-activation value for $b$ to equal $s$ when the input is reflected across the vertical axis, therefore that:

$ \sum\limits{_{i=0}^{n - 1}} \sum\limits{_{j=0}^{m - 1}} {a_{ij} x_{ij}} = \sum\limits{_{i=0}^{n - 1}} \sum\limits{_{j=0}^{m - 1}} {b_{ij} x_{i{(m - j)}}}$

For any $x$. Thus we need: $a_{ij} = b_{i{(m-j)}}$ for any $i$, $j$.

This is demonstrated in Figure~\ref{fc_invar_graphic}. Again a similar proof can be built for vertical and diagonal reflections, and can be applied to the biases of each layer. These tactics also readily generalize to cases where there are multiple channels in the input and output. Encoding each reflection invariance ties each weight to another weight, so when accounting for all three possible reflections each each weight is tied to $2\times2\times2 = 8$ other weights. Likewise the weight tying in the convolutional filters reduced the number of parameters in each filter by approximately an eighth. Thus we have reduced the number of parameters in the network by about eight folds. Averaging the weights requires a bit of computation, but in our experience it is a very minor cost relative to forward and backward propagating batches of training data.

An alternative method of enforcing horizontal reflection preservation would be to apply the filters to half the input, and then to reflect those filters and apply them to the other half of the input. This would allow meeting the requirement that $w_{ij} = w_{i(m -j)}$ without tying the weights, but we have not experimented with this approach.

While designed for Go, techniques in this vein have a ready application to image processing where we often expect the target function to be invariant to horizontal reflections. With a minor adjustment one could make a layer that is invariant to reflections rather then reflection preserving, meaning if its input is reflected its output will not change. Referring to Figure ~\ref{fc_invar_graphic}, this is just a matter of ensuring $b_{02} = b_{00}$, $b_{10} = b_{12}$, and $b_{20} = b_{22}$. Then one could build a network where all the lower layers preserve horizontal reflections and the final layer is invariant to horizontal reflections. The resulting network will be invariant to horizontal reflections, and have half the number of parameters of an untied network. This provides a way to account for the expected reflectional invariance property without having to double the amount of training data.

\section{Training}
\subsection{Datasets}
We test our network on two datasets, The first is the Games of Go on Disk\footnote{http://gogodonline.co.uk/} (GoGoD) dataset consisting of 81,000 professional Go games. Games are played under a variety of rulesets (but usually Chinese rules) and have long time controls. The second dataset consists of 86,000 Go games played by high ranked players on the KGS Go server~\footnote{https://www.gokgs.com/}\footnote{http://u-go.net/gamerecords/}. These games are all played under Japanese rules, have slightly lower player rankings, and generally use much faster time controls. We use two datasets because previous work in this field has typically used either one or the other. We use all available data from the GoGoD dataset, we select 86,000 games for the KGS dataset (where games are on average slightly shorter) so that the number of position-moves pairs in each dataset that can be trained on is roughly equal. This yields about 16.5 million move-positions pairs for each dataset.

\subsection{Methodology}
We partition the datasets into test, train, and validation sets each containing position-move pairs that are from disjoint games of Go. We use 8\% (1.3 million) for testing, 4\% (620 thousand) for validation, and the remaining (14.7 million) for training. We use the validation set to monitor learning and to experiment with hyperparameters. We found that vanilla gradient descent was effective for this task, although we found that it was important to anneal the learning rate towards the end of learning. We do not use dropout~\cite{hinton2012droppout}, in part because doing so would increase training time and in part because overfitting is not our primarily problem. We also do not use l2 or l1 regularization. Both convolutional and fully connected layers have their biases initialized to zero and weights drawn from a normal distribution with mean 0 and standard deviation 0.01.
\section{Results}
\subsection{Leave One Out Analysis}
\label{loo_section}
\begin{table}[ht]
\centering
\begin{tabular}{| c | c | c | c |}
\hline
Excluding & Accuracy & Rank & Probability \\ \hline
None &36.77\% & 7.50 & 0.0866 \\ \hline
Ko Constraints & 36.55\% & 7.59 & 0.0853 \\ \hline
Edge Encoding & 36.81\% & 7.64 & 0.0850 \\ \hline
Masked Training & 36.31\% & 7.66 & 0.0843 \\ \hline
Liberties Encoding & 35.65\% & 7.89 & 0.0811 \\ \hline
Reflection Preserving & 34.95\% & 8.32 & 0.0760 \\ \hline
All but Liberties & 34.48\% & 8.36 & 0.0755\\ \hline
All & 33.45\% & 8.76 & 0.0707 \\ \hline
\end{tabular}
\caption{Ablation Study. Here we train a medium scale network consisting for four convolutional layers and one fully connected layer on the GoGoD dataset while excluding different features.  Networks are judged based on their accuracy, average probability they assign the correct move, and average rank they give the correct move relative to the other possible moves on the test set. The liberty encoding and reflection preserving techniques are the most useful, but all the suggested techniques improve average rank and average probability.}
\label{loo_table}
\end{table}

To test some of the design choices made here we performed an ablation study. The study was done on a `medium scale' network to allow multiple experiments to be conducted quickly. The network had one convolutional layer with 48 7x7 filters, three convolutional layers with 32 5x5 filters, and one fully connected layer. Networks were trained with mini-batch gradient descent with a batch size of 128, using a learning rate of 0.05 for 7 epochs, and 0.01 for 2 epochs which took about a day on a Nvidia GTX 780 GPU. The results are in Table~\ref{loo_table}. The reflection preserving technique was extremely effective, leaving it out dropped accuracy by almost 2\%. The liberty encoding scheme improved performance but was not as essential, leaving it out dropped performance by 1\%. The remaining optimizations had a less dramatic impact but still contributed non-trivially. Together these additions increased the overall accuracy by over 3\% and increased accuracy relative to just using liberties encoding by over 2\%.
\subsection{Full Scale Network}

\begin{figure}[ht]
\begin{center}
\includegraphics[scale=0.8]{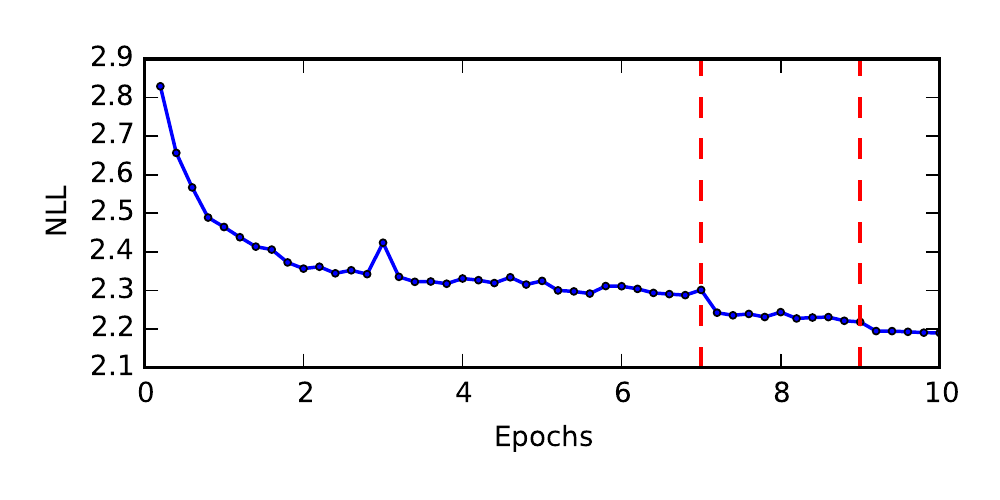}
\caption{Negative log likelihood on the validation set while training the 8 layer DCNN trained on the GoGoD dataset. Vertical lines indicate when the learning rate was annealed. Improvement on the validation set has more or less slowed to a halt despite only using 10 epochs of training.}
\label{final_network_learning_rates}
\end{center}
\end{figure}

\begin{table}[ht]
\centering
\begin{tabular}{| c | c | c | c |}
\hline
 & Accuracy & Rank & Probability \\ \hline
Test Data & 41.06\% & 5.91 & 0.1117 \\ \hline
Train Data & 41.86\% & 5.78 & 0.1158 \\ \hline
\end{tabular}
\caption{Results for the 8 layer DCNN on the train and test set of the GoGoD dataset. Rank refers to the average rank the expert's move was given, Probability refers to the average probability assigned to the expert's move.}
\label{gogod_results}
\end{table}

\begin{table}[ht]
\centering
\begin{tabular}{| c | c | c | c |}
\hline
 & Accuracy & Rank & Probability \\ \hline
Test Data & 44.37\% & 5.21 & 0.1312 \\ \hline
Train Data & 45.24\% & 5.07 & 0.1367 \\ \hline
\end{tabular}
\caption{Results for the 8 layer DCNN on the train and test set of the KGS dataset. Rank refers to the average rank the expert's move was given, Probability refers to the average probability assigned to the expert's move}
\label{kgs_results}
\end{table}

\begin{figure}[ht]
\begin{center}
\includegraphics[scale=0.8]{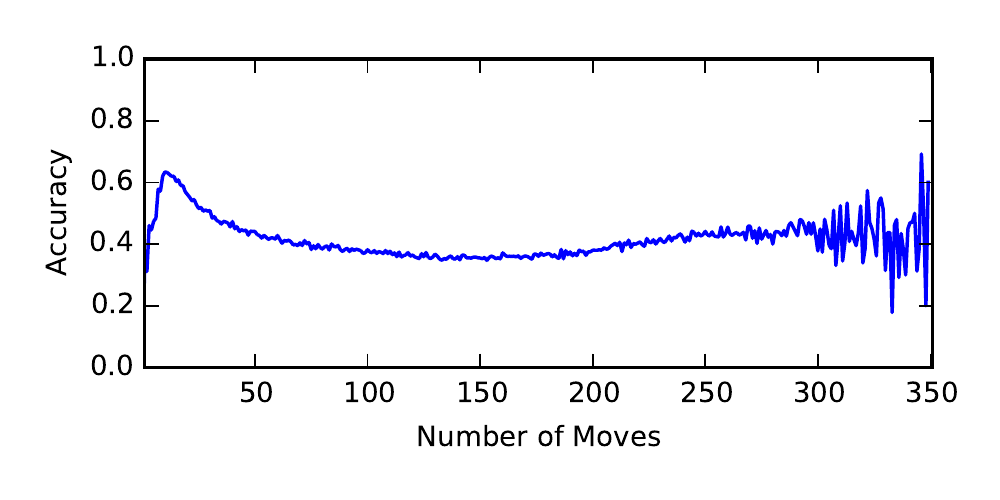}
\caption{Accuracy when a fixed number of moves have passed on the GoGoD test set. The network is more accurate during the beginning and end game and less accurate during the middle game.}
\label{accuracy_vs_moves}
\end{center}
\end{figure}

\begin{figure}[ht]
\begin{center}
\includegraphics[scale=0.8]{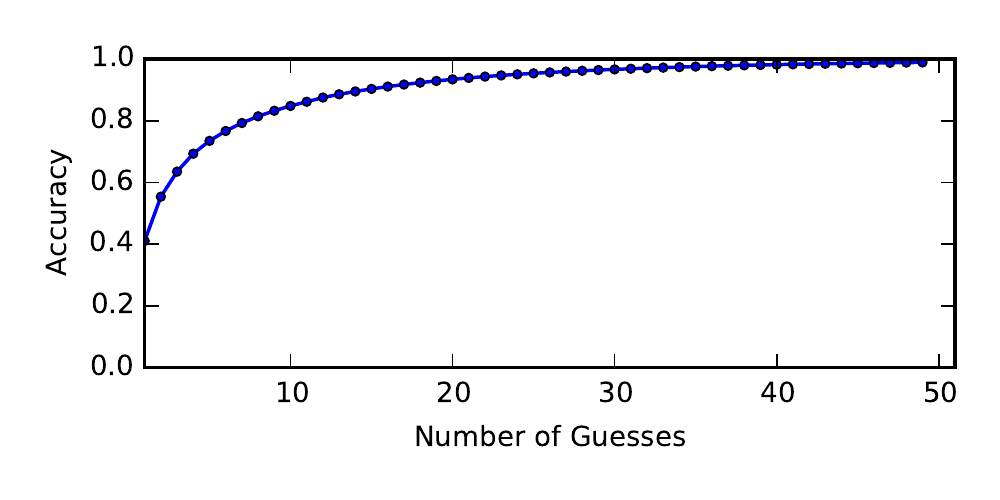}
\caption{Accuracy when allowed a fixed number of guesses on the GoGoD test set. The sharp increase in accuracy that occurs when given a few additional guesses indicates that, if the neural network's best guess is wrong, the right answer is often among its next few guesses. However there are occasions when the right move is ranked as the 30-50th best move by the network.}
\label{accuracy_vs_guess}
\end{center}
\end{figure}

The best network had one convolutional layer with 64 7x7 filters, two convolutional layers with 64 5x5 filters, two layers with 48 5x5 filters, two layers with 32 5x5 filters, and one fully connected layer. The network used all the optimizations enumerated in the previous section. The network was trained for seven epochs at a learning rate of 0.05, two epochs at 0.01, and one epoch at 0.005 with a batch size of 128 which took roughly four days on a single Nvidia GTX 780 GPU. Figure~\ref{final_network_learning_rates} shows the learning speed as measured on the validation set. The network was trained and evaluated on the GoGoD and KGS dataset, as shown in Table~\ref{gogod_results} and Table~\ref{kgs_results} respectively. To our knowledge the best reported result on the GoGoD dataset is 36.9\% using an ensemble of shallow networks~\cite{sutskever2008mimicking_go_experts} and the best on the KGS dataset is 40.9\% using feature engineering and latent factor ranking~\cite{wistuba2013move_prediction_modeling_feature_interaction}. Our results were completed on more recent versions of these datasets, but in so much as they can be compared our results has surpassed both these results by margins of 4.16\% and 3.47\% respectively.  Additionally, this was done without using the previous moves as input.~\cite{wistuba2013move_prediction_modeling_feature_interaction} did not analyze the impact using this feature had, but~\cite{sutskever2008mimicking_go_experts} report the accuracy of one of their networks dropped from 34.6\% to 21.8\% when this feature was removed, making it seem like their networks relied heavily on this feature.

We also examine the GoGoD test set accuracy of the network as a function of the number of moves completed in the game, see Figure~\ref{accuracy_vs_moves}. Accuracy increases during the more predictable opening moves, decreases during the complex middle game, and increases again as the board fills up and the number of possible moves decreases towards the end game. Finally, we examine how accurate the network is when allowing the network to make multiple guesses on the same test set, see Figure~\ref{accuracy_vs_guess}. It is encouraging to note that, if the network's highest ranking move is incorrect, its second or third highest ranked move is often correct. However there are times when the expert move was not among the top 40 ranked moves from the network. While it is not clear exactly how well a human expert would perform on this task, it seems likely that a human expert would practically always be able to guess where another expert would move given 40 guesses. Thus we do not think our DCNN has reached a human level of performance on this task.

\subsection{Playing Go}
\begin{table}
\centering
\begin{tabular}{| c | c | c |}
\hline
Network & Chinese Rules & Japanese Rules \\ \hline
KGS & 0.86 & 0.85 \\ \hline
GoGoD & 0.87 & 0.91 \\ \hline
GoGoD Small & 0.71 & 0.67 \\ \hline
\end{tabular}
\caption{Win rates of three DCNNs against GNU Go level 10 using Chinese and Japanese rules. For each matchup 200 games were played. GoGoD and KGS refer to the full scale network trained on the named dataset, GoGoD Small refers to the exclude none network from Section~\ref{loo_section}. Even the smaller DCNN is able to consistently defeat GNU Go.}
\label{gnugo_table}
\end{table}

\begin{table}
\centering
\begin{tabular}{| c | c |}
\hline
Network & Win Rate \\ \hline
KGS & 0.12 \\ \hline
GoGoD & 0.14 \\ \hline
\end{tabular}
\caption{Winrates when playing the 8 layer DCNN trained on the GoGoD and KGS dataset against Fuego 1.1 with Chinese rules. Fuego was given 10 seconds a move, four gigabytes of RAM, and 2 threads on an Intel i7-4702MQ processor. Pondering (performing computations during the opponent's turn) was turned off. For each matchup 200 games are played. Being able to win even a few games against this opponent indicates a high degree of skill was acquired.}
\label{fuego_table}
\end{table}

The networks trained here were successful move predictors, but that does not necessary mean they will be strong Go players. There are two potential problems. First, during a game an opponent, or the classifier itself, are liable to make moves that create positions that are uncommon for games between professionals. Since the networks have not been tested or trained on these kinds of positions there is no guarantee they will continue to perform well when this occurs. Second, even if the classifier is able to predict an expert player level move 50\% of the time, if its other moves are extremely poor it could still be a terrible Go player. To test the strength of the networks as Go players they were played against two other computer Go programs, GNU Go 3.8\footnote{https://www.gnu.org/software/gnugo/} and Feugo 1.1~\cite{fuego}. We test the final network trained on the KGS data, the GoGoD data, and the smaller `exclude none' network from Section~\ref{loo_section}. The results can be found in Table~\ref{gnugo_table} and Table~\ref{fuego_table} for GNU GO and Fuego respectively. There is one complication, the DCNNs do not have the capability to pass during their turn. Therefore, should a game go on indefinitely, the networks will eventually run out of good moves to play and start making suicidal moves thus losing the game. To work around this issue we allow both Fuego and GNU Go to resign, we additionally have the DCNN pass its turn whenever its opponent does thus ending the game. Games were scored using the opposing Go engine's scoring function.

The results are very promising. Even though the networks are playing using a `zero step look ahead' policy, and using a fraction of the computation time as their opponents, they are still able to play better then GNU Go and take some games away from Fuego. Under these settings GNU Go might play at around a 6-8 kyu ranking and Fuego at 2-3 kyu, which implies the networks are achieving a ranking of approximately 4-5 kyu. For a human player reaching this ranking would normally require years of study. This indicates that sophisticated knowledge of the game was acquired. This also indicates great potential for a Go program that integrates the information produced by such a network. The smaller network we trained comfortably defeated GNU Go despite being less accurate then some previous work at move prediction. Thus it seems likely that our choice not to use the previous move as input has helped our move predictors to generalize well from predicting expert Go player's moves to playing Go as stand alone players. This might also be attributed to our deep learning based approach. Deep learning algorithms have been shown in particular to benefit from out of sample distributions~\cite{bengio2011deep_out_of_dist}, which relates to our situation since our networks can be viewed as learning to play Go from a biased sample of positions. The network trained on the GoGoD dataset performed slightly better than the one trained on the KGS dataset. This is what we might expect since the KGS dataset contains many games of speed Go that are liable to be of lower quality.

\section{Conclusion and Future Work}
In this work we have introduced the application of deep learning to the problem of predicting the moves made by expert Go players. Our contributions also include a number of techniques that were helpful for this task, including a powerful weight tying technique to take advantage of the symmetries we expect to exist in the target function. Our networks are state of the art at move prediction, despite not using the previous moves as input, and can play with an impressive amount skill even though future positions are not explicitly examined.

There is a great deal that could be done to extend this work. Scaling up is likely to be effective. We limited the size of our networks to keep training time low, but using more data or larger networks will likely increase accuracy. We have only completed a preliminary exploration of the hyperparameter space and think better network architectures could be found. Curriculum learning~\cite{bengio2009curriculum} and integration with reinforcement learning techniques might provide avenues for improvement. The excellent skill achieved with minimal computation could allow this approach to be used for a strong but fast Go playing mobile application. The most obvious next step is to integrate a DCNN into a full fledged Go playing system. For example, a DCNN could be run on a GPU in parallel with a MCTS Go program and be used to provide high quality priors for what the strongest moves to consider are. Such a system would both be the first to bring sophisticated pattern recognitions abilities to playing Go, and have a strong potential ability to surpass current computer Go programs.

\fontsize{9.5}{10.5}
\selectfont
\bibliography{paper}
\bibliographystyle{icml2014}

\end{document}